\documentclass[11pt]{article}
\usepackage[dvips]{graphicx}
\newcommand{\oo}{\bot}            
\newcommand{\pp}{\top}            
               
\newcommand{\legal}[2]{\mbox{\bf Lr}^{#1}_{#2}} 
\newcommand{\win}[2]{\mbox{\bf Wn}^{#1}_{#2}} 
\newcommand{\watom}[2]{#1\{#2\}}

\newcommand{\gneg}{\mbox{\small $\neg$}}                  
\newcommand{\mli}{\hspace{2pt}\mbox{\small $\rightarrow$}\hspace{2pt}}                      
\newcommand{\mld}{\hspace{2pt}\mbox{\small $\vee$}\hspace{2pt}}     
\newcommand{\mlc}{\hspace{2pt}\mbox{\small $\wedge$}\hspace{2pt}}   
\newcommand{\add}{\hspace{2pt}\mbox{\small $\sqcup$}\hspace{2pt}}                     
\newcommand{\adc}{\hspace{2pt}\mbox{\small $\sqcap$}\hspace{2pt}} 
\newcommand{\tlg}{\bot}               
\newcommand{\twg}{\top}               

\newtheorem{theoremm}{Theorem}[section]
\newtheorem{factt}[theoremm]{Fact}
\newtheorem{corollaryy}[theoremm]{Corollary}
\newtheorem{definitionn}[theoremm]{Definition}
\newtheorem{thesiss}[theoremm]{Thesis}
\newtheorem{lemmaa}[theoremm]{Lemma}
\newtheorem{conventionn}[theoremm]{Convention}
\newtheorem{examplee}[theoremm]{Example}
\newtheorem{exercisee}[theoremm]{Exercise}
\newtheorem{remarkk}[theoremm]{Remark}
\newenvironment{definition}{\begin{definitionn} \em}{ \end{definitionn}}

\newcommand{\propge}{\mbox{\bf CL2}}
\newcommand{\propgew}{\mbox{\bf CL2$^{\Psi}$}}

\newcommand{\propgewo}{\mbox{\bf CL2$^{o,\Psi}$}}
\newcommand{\propel}{\mbox{\bf CL1}}

\newcommand{\propeltw}{\mbox{\bf CL12}}
\newcommand{\nheu}[1]{ \bar{#1}}   

\begin{document}
\begin{center}
{\Large {\bf Implementing Agent-Based Systems via Computability Logic \propge}}
\\[20pt] 
{\bf Keehang Kwon}\\
Dept. of Computer Engineering, DongA University \\
Busan 604-714, Korea\\
khkwon@dau.ac.kr
\end{center}

{\em Computability logic} (CoL) is a powerful computational model.
In this paper,  we show that CoL
naturally supports   multi-agent
programming models where resources (coffee for example) are involved. 
To be specific, we discuss an implementation of the Starbucks
 based on CoL (\propge\ to be exact).


\section{Introduction}\label{sintr}

The design and  implementation  of multi-agent systems
is recognized  as a key component of general AI. Yet it remains the case that existing approaches -- classical logic,
$\pi$-calculus, linear logic, etc -- are too simplistic to
 encode real-world multi-agent systems. Implementing the Starbucks
in AI is such an example.

{\em Computability logic} (CoL) \cite{Jap0}-\cite{JapCL12}, is an
elegant theory of (multi-)agent computability. In CoL, computational problems are seen as games between a machine and its environment and logical operators stand for operations on games.
It understands interaction among agents in its most general --- game-based --- sense. 
There are many fragments of CoL.
To represent resources such as coffee, we choose \propge -- a basic fragment of CoL -- as our target language.  \propge\ is  obtained  by adding to    \propel\ a second kind of atoms called
{\it general atoms}. A general atom  
models an arbitrary interactive computing problem such as a coffee
machine.

 In this paper, we discuss a web-based implementation of multi-agent programming  based on \propge \cite{JapCL2}.
 We assume the following in our model:

\begin{itemize}

\item Each agent corresponds  to a web site with a URL. An agent's  resourcebase(RB) is described in its homepage.


\item There are three kinds of agents: God, resource providers/consumers  and regular agents.
  A resource provider for a resource $R$, written as $*R$, is an agent who is given a resource manual  by God.
    It  can thus produce as many copies of resource $R$ using the manual. 
    A resource consumer is an agent who gives the  resource to  God.

\item   God is   both the ultimate provider for every
  resource and its ultimate consumer.  
  
  \item Our goal here is to program every agent including resource providers/consumers. For this, we assume that 
a resource provider has a  machine's manual/heuristic  $h$ for creating a resource R.  
 Similarly, the counterstrategy of a resource consumer 
- the consumer's script -- is  preprogrammed in the environment's strategy $\nheu{s}$.  Note that, unlike the machine's 
strategy, a consumer's counterstrategy varies from a resource customer to a resource customer.
To represent these manuals/scripts, we extends general atom P to $\watom{P}{h}$  for  resource providers,
general atom P to $\watom{P}{\nheu{s}}$  for resource consumers. 

\end{itemize}

In this paper, we present \propgew\
which is a  web-based implementation of \propge. This implementation is rather  simple and
straightfoward.
What is interesting is that \propgew\ is an appealing multi-agent
programming model where resources are involved.

\section{Preliminary}

We review the basic relevant concepts of CoL, and some basic notational conventions.  A reader may want to consult \cite{JapCL1} for further details. 


 CoL    understands the interactive computational problems as games between two players:  {\em machine} and {\em environment}. The symbolic names   for these two players are   
$\twg$ and 
$\tlg$,
 respectively.

A
{\bf move} means  a finite string over the  keyboard alphabet. 
A  {\bf labmove} is a move prefixed with $\top$ or $\bot$.  A {\bf run} is a (finite or infinite) sequence of labmoves, and a 
{\bf position} is a finite run.
Runs will be  delimited by ``$\langle$'' and ``$\rangle$''.
 $\langle\rangle$  denotes the {\bf empty run}.

The following is a brief definition of the concept of a constant game.

\begin{definition}\label{game}
 A {\bf constant game}  is a pair $A=(\legal{A}{},\win{A}{})$, where:\vspace{10pt}

1. $\legal{A}{}$ is a set of runs satisfying the condition that a finite or infinite run is in $\legal{A}{}$ iff all of its nonempty finite --- not necessarily proper --- initial
segments are in $\legal{A}{}$ (notice that this implies $\langle\rangle\in\legal{A}{}$). The elements of $\legal{A}{}$ are
said to be {\bf legal runs} of $A$. \vspace{5pt} 

2. $\win{A}{}$ is a function that sends every run $\Gamma$ to one of the players $\top$ or $\bot$. 
\end{definition}

Unfortunately, the above definition is not sufficient
to represent  games equipped with some kind of
heuristics.   AlphaGo is such an example.
For this reason, we introduce a new game which we call 
{\it a constant game with  heuristics}, denoted by
$\watom{A}{h}$ where $h$ is a heuristic function.
For example,  AlphaGo can be represented by $\watom{Go}{h}$ where
$h$ represents the powerful heuristics of the AlphaGo.

\begin{definition}\label{game}
 A {\bf constant game with heuristics}  is a pair $\watom{A}{h} =(\legal{A}{},\win{A}{},h^A)$, where:\vspace{10pt}

 $h^A$ is a heuristic function for the machine to follow, i.e., the machine's strategy for the game $A$. 
  $h^A$ typically depends on the run of the game.
\end{definition}

Often we need to preprogram the environment's strategy as well.
For this reason, we introduce a new game which we call 
{\it a constant game with  environment's strategy }, denoted by
$\watom{A}{\nheu{s}}$ where $\nheu{s}$ describes the environment's strategy for the game.

\section{\propgew}\label{s2tb}

 We  review the  propositional computability 
 logic  called $\propge$ \cite{JapCL1}.  
 
 As always, there are infinitely many {\bf elementary atoms} in the language, for which we will be using the letters
 $p,q,r,\ldots$ as metavariables. There are also infinitely many
 {\bf general atoms} in the language, for which we will be using the letters $P,Q,R,\ldots$.
  We introduce {\bf general atoms with machine's strategy/heuristics}, denoted by $\watom{P}{h},\ldots$ and
  {\bf general atoms with environment's strategy}, denoted by $\watom{P}{\nheu{s}}, \ldots$.
  
  The two  atoms: $\twg$ and $\tlg$ have a special status in that their interpretation is fixed.  Formulas of this language, referred to as {\bf $\propge$-formulas}, are built from atoms in the standard way:

\begin{definition}
  
  The class of $\propge$-formulas 
is defined as the smallest set of expressions such that all atoms are in it and, if $F$ and $G$ are in it, then so are 
$\gneg F$, $F\mlc G$, $F \mld G$, $F \mli G$, $F\adc G$, $F \add G$. 
\end{definition}

Now we define $\propgew$, a slight extension to $\propge$ with environment parameters.
Let $F$ be a  $\propge$-formula.
We introduce a new {\it env-annotated} formula $F^\omega$ which reads as `play $F$ against an agent $\omega$.
For an $\adc$-occurrence $O$ (or an occurrence $O$ of a general atom) in $F^\omega$, we say
$\omega$ is the {\it matching} environment of $O$.
For example, $(p \adc (q \adc r))^{w}$ is an  agent-annotated formula and
$w$ is the matching environment of both occurrences of $\adc$.
We extend this definition to subformulas and formulas. For a subformula $F'$ of the above $F^\omega$,
we say that $\omega$ is the {\it matching} environment of both $F'$ and $F$.

In introducing environments to a formula $F$, one issue is
whether we allow `env-switching' formulas of the form
$(F[R^u])^w$. Here $F[R]$ represents a formula with some occurrence of a subformula $R$.
That is, the machine initially plays $F$ against agent $w$ and then switches
to play against another agent $u$ in the course of playing $F$.
For technical reasons,  we focus on non `env-switching' formulas.
This leads to the following definition where $h$ is a heuristic function:

\begin{definition} 
  The class of $\propgew$-formulas
is defined as the smallest set of expressions such that
(a) For any $\propge$-formula $F$ and any agent $\omega$, $F^\omega$  are in it and, (b) if $H$ and $J$ are in it, then so are 
$\gneg H$, $H\mlc J$, $H\mld J$, $H\mli J$.
\end{definition}

\begin{definition}
\noindent Given a $\propgew$-formula $J$,  the skeleton  of $J$ -- denoted by
$skeleton(J)$ -- is obtained by
replacing every occurrence $F^\omega$ by $F$.
\end{definition}
\noindent For example, $skeleton((p \adc (q \adc r))^{w}) = p \adc (q \adc r)$.

We often use $F$ instead of $F^{\omega}$ when it is irrelevant.
.

The following definitions comes from \cite{JapCL2}. They  apply both
to $\propge$,  and $\propgew$.

Understanding $E\mli F$ as an abbreviation of $\neg E \mld F$, a {\bf positive} occurrence of a subformula is one that is in the scope of an even number of $\neg$'s. Otherwise, the occurrence is {\bf negative}.

A {\bf surface occurrence} of a subformula means an occurrence that is not in the scope of a choice ($\add$ or $\adc$) operator.

A formula is  {\bf elementary} iff it does not contain the choice operators and general atoms.

The {\bf elementarization} of a formula is the result of replacing, in it, every surface occurrence of the form $F_1\add ... \add F_n$ by $\oo$ ,  every surface occurrence of the form $F_1\adc ... \adc F_n$ by $\pp$,
every positive surface occurrence of each general atom by $\oo$ , and
every negative
surface occurrence of the form  each general atom by $\pp$.

A formula is {\bf stable} iff its elementarization is valid in classical logic, otherwise
it is {\bf instable}.

$F${\bf -specification} of $O$, where  $F$ is a formula and $O$ is a surface occurrence in $F$, is a string $\alpha$ which can be defined by:
\begin{itemize}

\item $F$-specification of the occurrence in itself is the empty string.

\item If $F$ = $\neg G$, then $F$-specification of an occurrence that happens to be in $G$ is the same as the $G$-specification of that occurrence.

\item If $F$ is $G_1\mlc ... \mlc G_n$, $G_1\mld ... \mld G_n$, or $G_1\mli G_2$, then $F$-specification of an occurrence that happens to be in $G_i$ is the string $i.\alpha$, where $\alpha$ is the $G_i$-specification of that occurrence.

\end{itemize}

The proof system of \propgew\ is identical to that $\propge$ and
has the following three rules, with $H$, $F$ standing for $\propgew$-formulas and $\vec H$ for a set
of $\propgew$-formulas: \\

Rule (A): ${\vec H}\vdash F$, where $F$ is stable and, whenever $F$ has a positive (resp. negative) surface occurrence of $G_1\adc ... \adc G_n$ (resp. $G_1\add ... \add G_n$) whose matching environment is $\omega$, for each i$\in \{1,...,n\}$, $\vec H$ contains the result of replacing in $F$ that occurrence by $G_i^\omega$.

Rule (B): $H\vdash F$, where $H$ is the result of replacing in $F$ a negative (resp. positive) surface occurrence of $G_1\adc ... \adc G_n$ (resp. $G_1\add ... \add G_n$) whose matching environment is $\omega$ by $G_i^\omega$ for some i$\in \{1,...,n\}$.

Rule (C): $F'\vdash F$, where $F'$ is the result of replacing in $F$
 one negative  surface occurrence of
some general atom $P$  and one positive  surface occurrence of
some general atom $P$ by a nonlogical elementary atom that does not occur
in $F$.

\begin{examplee}\label{ex01}

$\propgew \vdash $$(C \mlc C)\mli (C\mld C)^\omega$

  where $\omega$ is an agent.
  Note that $\omega$ play no roles in the proof procedure. Similarly, the machine's manual and the environment's script
  play no role in the proof procedure.
\end{examplee}
\begin{enumerate}

\item $(p\mlc q)\mli (p\mld q)^\omega$, rule A, 0

\item $(p\mlc  C) \mli (p\mld C)^\omega$, rule C, 1

\item $(C \mlc C)\mli (C\mld C)^\omega$, rule C, 2

\end{enumerate}

\section{Hyperformulas}\label{s3hyper}

To facilitate the execution procedure, we modify \propgew\ to obtain
\propgewo.
Unlike \propgew, this new language allows any  hyperformulas.
Its rules 
are Rules (a) and (b) of \propgew\ plus the following Rule (c$^o$)
instead of the old Rule (c): \\

Rule (C$^o$): $F'\vdash F$, where $F'$ is the result of replacing in $F$ one negative  surface occurrence of
some general atom $P$  and
 one positive   surface occurrence of
some general atom $P$ by a hybrid atom $P_q$.

In the above, we introduced hybrid atoms. Each  hybrid atom is a pair consisting of a general atom $P$, called its general component, and a nonlogical elementary atom $q$, called its elementary component. Hybrid atoms were introduced
in \cite{JapCL2} to distinguish elementary atoms introduced in Rule (c)
 from all other elementary atoms. 
 
 Now atoms can be of one of the three (elementary, general or hybrid)
 sorts. 
All the terminologies and definitions of the previous section extends well to hyperformulas.
 One exception is that in the  elementarization
of a hyperformula, every surface occurrence of each hybrid atom must also be replaced
by the elementary component of that atom. 
%

We can easily convert $\propgew$ proof to a modified one: if $q$ is obtained from $P$ by Rule (c), replace all 
occurrences of $q$ by $P_q$.  Apply this procedure to all of its descendants in the proof tree as well.

\begin{examplee}\label{ex01}

$\propgewo \vdash (C \mlc C)\mli (C\mld C)^\omega$

  where  $\omega$ is an agent.
  Note that $\omega$ play no roles in the proof procedure.
\end{examplee}
\begin{enumerate}

\item $(C_p \mlc C_q)\mli (C_p\mld C_q)^\omega$, rule A, 0

\item $(C_p\mlc  C) \mli (C_p\mld C)^\omega$, rule C, 1

\item $(C \mlc C )\mli (C\mld C)^\omega$, rule C, 2

\end{enumerate}

\section{Execution Phase}\label{s22tog}

The machine model of \propge\ is designed to process only one query/formula at one time.
In distributed systems such as \propgew, however, it is natural for an agent to receive/process
multiple queries. For this reason, our machine processes multiple formulas one by one.

Multiple queries cause some complications, as the RB of the machine evolves to RB' in the course of solving a query.
In such a case, subsequent queries must be solved with respect to RB'.
To be specific, it maintains a queue  $Q = \langle Q_1,\ldots,Q_n \rangle$ for storing multiple incoming
queries. We assume that the machine processes $Q_1,\ldots,Q_n$ by executing the following $n$
 procedures sequentially:

\[ Exec(RB_1\mli Q_1), Exec(RB_1\mli Q_2)\ldots, Exec(RB_n\mli Q_n) \]
\noindent
Here $RB_1$ is the original RB associated with the machine.  We assume here that, for
$1\leq i\leq n$, $RB_i$ evolves
to $RB_{i+1}$ after solving $Q_i$.

It leads to the following definition: \\\\

{\bf procedure} EXEC(K,Q):   $K$ is RB of the agent and $Q$ is a queue of incoming queries.\\

\begin{itemize}

\item If $K =  \Gamma$ and $Q = (Q_1,\ldots,Q_n)$ then we  do the following:

In this case, the machine tries to solve the first 
query  by invoking  $Exec(\Gamma\mli Q_1)$ and then EXEC$(\Gamma', (Q_2,\ldots,Q_n)) $.

\item Else ($Q$ is empty): wait for  new incoming service calls. 

\end{itemize}

Below we will introduce an algorithm that executes
a formula $J$. The algorithm is a minor variant
of the one in \cite{JapCL2} and contains two stages: \\

{\bf Algorithm Exec(J)}: \%  $J$ is a $\propgew$-formula \\

\begin{enumerate}

\item First stage is to initialize a temporary variable $E$ to $J$,
 a position variable $\Omega$ to an empty position $\langle\rangle$.
Activate all the  agents specified in $J$.
          
\item The second stage is to play $J$ according to the following $mainloop$ procedure
(which is from \cite{JapCL2}): 

\end{enumerate}

procedure $mainloop(Tree)$: \%  $Tree$ is a proof tree of $J$ \\

 {\bf Case} $E$ is derived by Rule (B): \\
  \hspace{3em}       Let $H$ be the premise of $E$ in the proof. $H$ is the result of substituting, in $E$, a certain negative (resp. positive) surface occurrence of a subformula $G_1\adc\ldots\adc G_n$ (resp. $G_1\add\ldots\add G_n$) by $G^\omega_i$ for some $i\in\{1,\ldots,n\}$.  Here, we assume that
  $\omega$ is the matching environment of that occurrence.
  Let $\gamma$ be the $E$-specification of that occurrence. Then make the move $\gamma i$, update $E$ to $H$.  Then inform $\omega$ of the move  $\gamma i$.
  repeat $mainloop$
  
 {\bf Case} $E$ is derived by Rule (C$^0$): \\
  \hspace{3em}  
  Let $H$ be the premise of $E$ in the proof. $H$ is the result of
  replacing in $E$ some positive surface occurrence $\pi$ and some
  negative surface occurrence $\nu$ of a general atom $P $ by a hybrid atom 
  $P_q $. Let $\langle\oo\pi_1,\ldots,\oo\pi_n\rangle$ and $\langle\oo\nu_1,\ldots,\oo\nu_m\rangle$
  be $\Omega^\pi$ and $\Omega^\nu$, respectively\footnote{$\Omega^\pi$ and $\Omega^\nu$ may be  programmed in
  $h$ in $\watom{P}{h}$.}.
  Here $\Omega^\pi$ is the subrun of the occurrence $\pi$ and $\Omega^\nu$ is the subrun of the occurrence $\nu$ of the  hybrid atom introduced.
  Then:
  make the $m+n$ moves $\pi\nu_1,\ldots,\pi\nu_m,\nu\pi_1,\ldots,\nu\pi_n$
  (in this  order); update $\Omega$ to $\langle\Omega,\pp\pi\nu_1,\ldots,
  \pp\pi\nu_m,\pp\nu\pi_1,\ldots,\pp\nu\pi_n\rangle$.
  Update E to H; repeat $mainloop$.
  
{\bf Case} $E$ is  derived by  Rule (a): \\

Follow the procedure innerloop described below.
Below, ``the environment makes a move'' means that
either the environment makes a move or $\pp$ makes a 
move for the environment using a given heuristic function.\\

$innerloop$: Keep granting permission until the environment
makes a move $\alpha$.  \\

 Subcase (i): $\alpha=\gamma\beta$, where $\gamma$
$E$-specifies a surface occurrence of a general atom. Then update $\Omega$
 to $\langle\Omega,\oo\gamma\beta\rangle$
 and repeat $innerloop$. \\
 
Subcase (ii): $\alpha=\gamma\beta$, where $\gamma$ $E$-specifies a surface occurrence of a hybrid atom. Let $\sigma$  be the $E$-
specification of the other occurrence of the same hybrid atom.
Then make the move $\sigma\beta$, update $\Omega$ to $\langle\Omega,\oo\gamma\beta,\pp\sigma\beta\rangle$ and repeat $innerloop$. \\

Subcase (iii): $\alpha  = \gamma i$, where $\gamma$ $E$-specifies a  positive (negative) surface occurrence of a
    subformula $G_1\adc\ldots\adc G_n$ ($G_1\add\ldots\add G_n$) and $i\in\{1,\ldots,n\}$. 
    Let   $H$ be the result of substituting, in $E$, a certain negative (resp. positive) surface occurrence of a subformula $G_1\add\ldots\add G_n$ (resp. $G_1\adc\ldots\adc G_n$) by $G^\omega_i$ for some $i\in\{1,\ldots,n\}$. Here $\omega$ is the matching environment of that occurrence.
    Then update $E$ to $H$, and repeat $mainloop$. \\
    
If $\alpha$ does not satisfy the conditions of any of the above Subcases (i),(ii),(iii), ignore it.



\section{Examples}\label{sec:modules}

As an example of multi-agent system, we will look at the Starbucks. This example introduces 
several interesting concepts such as how service $flows$ among agents.
It is formulated with the God, the Folger coffee maker (coffee provider), the Starbucks owner, a user and the bank (dollar provider).
We assume the following:

\begin{itemize}

\item  
In our example, God
provides the coffee-making manual to the Folger and collects \$10. It also provides  the dollar-making manual
to the bank and collects ten coffees.

\item God is not actually implemented. Instead, the Folger and the bank play the role of God whenever necessary.

\item The store owner  plays the roles of barista and cashier.

\item The owner  tries to borrow \$8 from the bank and pay 8 coffees to it. He also tries
to pay \$10 and gets  ten coffees from the Folger.
    
\item Each coffee costs a dollar.
  
\item The user tries to get two coffees by paying two dollars to the owner.
    He also tries to get two dollars by paying two coffees to the bank.
    
 \item The user is  active from the beginning.

\end{itemize}  
  
Now we want to implement the above.
The first task is to determine the representation of a coffee.
A coffee is represented by a (imaginary or real, depending on your
need) coffee machine.
We assume that the owner has a coffee manual/heuristic which provides the 'rules of thumb' to make a good coffee.

A coffee machine -- similar to an ATM machine --
can be seen as a game between the owner with a manual  and
its customer with a sequence of interactions.
Assume we have a particular coffee machine with LCD monitor where

(1) The user of the machine selects the $x (= 1,2,\ldots,)$ grams of sugar, and then
the $y (= 1,2,\ldots)$*10cc of milk

(2) The owner  selects the $z(= 1,2,\ldots,10)$ spoons of coffee.

For simplicity, we assume that the owner  uses the following  heuristic evaluating function

\[ h(x,y,z) = | z - xy -1 |. \]   
\noindent 
In other words, if it selects $z$ such that $z = xy+1$, then
it knows that he/she makes good coffees.

One simplest way of representing this device is to represent it as a general
atom $\watom{C}{h}$ with the above heuristic $h$\footnote{Coffee machine can be represented without 
using general atoms but it is cumbersome.}.

Similarly, the consumer's preference in coffee  can be programmed in the user's scripts.
Below illustrates some user's scripts $\nheu{c_0},\nheu{c_1}$ used in the example below in coffee making. \\
 
$\nheu{c_0} = \{\langle \oo 3,\oo 1\rangle \}$. \% 3 grams of sugar, 10cc of milk\\

$\nheu{c_1} = \{\langle \oo 4,\oo 2\rangle \}$. \%  4 grams of sugar, 20cc of milk\\

 



As in the case of coffee, the same approach can be employed to
represent a dollar, i.e., as a credit-card paying machine or a POS machine.
A credit-card paying machine can be seen as an interactive
constant game. To make things simple, we assume the bank is a  provider for one dollar and   $r$ is
a manual for making a dollar.

\newenvironment{exmple}{
 \begingroup \begin{tabbing} \hspace{2em}\= \hspace{3em}\= \hspace{3em}\=
\hspace{3em}\= \hspace{3em}\= \hspace{3em}\= \kill}{
 \end{tabbing}\endgroup}
\newenvironment{example2}{
 \begingroup \begin{tabbing} \hspace{8em}\= \hspace{2em}\= \hspace{2em}\=
\hspace{10em}\= \hspace{2em}\= \hspace{2em}\= \hspace{2em}\= \kill}{
 \end{tabbing}\endgroup}

 An example  is provided by the following  $*C,o,u,*1$ agents.
In $\watom{1}{\nheu{d_0}}$ of the *C agent, $\nheu{d}$ describes a preprogrammed God's 
requirements  in making the first dollar.
Similarly, in $\watom{C}{\nheu{c_0}}$ of the bank agent, $\nheu{c_0}$ describes a preprogrammed  
requests in making the first
 coffee.

Now consider  $\watom{C}{h}$ in *C. Here $h$  is a heuristic function for making a coffee.
That is, $h$ is a coffee-making manual.

\begin{exmple}
\> $agent\ *C$. \%  Folger coffee provider \\
\> $\nheu{d_0} = \ldots$ \% God's requirements in the first dollar \\
\> \vdots \\
\> $\nheu{d_9} = \ldots$ \%  God's  requirements in the tenth dollar\\
\> $h(x,y,z) = \ldots$ \% coffee-making manual \\
\> $ (( \watom{1}{\nheu{d_0}} \mlc \ldots \mlc \watom{1}{\nheu{d_9}}) \mli \watom{C}{h})^{God}$. \%  the coffee manual 
costs ten (customized) dollars.\\
\end{exmple}

\begin{exmple}
\> $agent\ o $. \% starbucks owner \\
\> $((C \mlc \ldots \mlc C) \mli (1 \mlc \ldots \mlc 1))^k$. \% pay 8 coffees and get \$8 from bank.\\
\> $( (1 \mlc \ldots \mlc 1) \mli (C \mlc \ldots \mlc C))^f$. \% pay \$10 and get 10 coffees from Folger.\\
\end{exmple}

\begin{exmple}
  \> $agent\ u$. \% the client\\  
  \> $((C \mlc  C) \mli (1 \mlc 1))^k$.  \% pay 2 coffees and get  \$2 from bank.\\
\> $((1 \mlc 1) \mli (C \mlc C))^o.$  \% pay two dollars and get two   coffees from owner. \\
\end{exmple}

\begin{exmple}
  \> $agent\ *1$. \% the bank\\  
  \> $\nheu{c_0} = \ldots$ \% God's requirements in the first coffee \\
\> \vdots \\
\> $\nheu{c_9} = \ldots$ \%  God's  requirements in the tenth coffee\\
\> $h(x,y,z) = \ldots$ \% coffee-making manual \\

  \> $r(\ldots) = \ldots$ \% dollar-making manual \\
 \> $((\watom{C}{\nheu{c_0}} \mlc \ldots \mlc \watom{C}{\nheu{c_9}}) \mli \watom{1}{r})^{God}$. \% dollar-making manual 
   costs 10  ( customized) coffees.\\
\end{exmple}

Now consider the user agent $u$. The user is active from the beginning and tries to do the following:
 (1) obtain two coffees from the owner and pass it along to the bank, and by (2) obtaining two dollars and then
 passing them along to the
owner. The task (2) easily succeeds, as $u$ makes two dollars by copying the moves of the bank 
(The bank makes moves according to the recipe $r$).
 From this, the agent 
$u$  successfully pays the owner $o$ two dollars. The owner $o$ pays \$10 to $*C$ (\$2 from $u$, \$8 from *1)
all using the copy-cat method.
 Upon request,  $*C$ makes  ten (real or imaginary) coffees using the
``coffee manual'' $h$.  $o$ makes ten coffees by copying $*C$.
Note that the user can make  two coffees and $*1$ can make 8 coffees both by copying $o$.

\section{Conclusion} \label{s5thr}

In this paper, we proposed a multi-agent programming model based on $\propgew$.
Unlike other formalisms such as LogicWeb\cite{Loke} and distributed logic programming\cite{LCF},
this model does not require any centralized control.
Our next goal is to replace $\propgew$ with much more expressive $\propeltw$\cite{Japtow}.

\bibliographystyle{plain}

\end{document}